\documentclass[a4paper,twoside]{article}

\usepackage{epsfig}
\usepackage{subcaption}
\usepackage{calc}
\usepackage{amssymb}
\usepackage{amstext}
\usepackage{amsmath}
\usepackage{amsthm}
\usepackage{multicol}
\usepackage{pslatex}
\usepackage{apalike}

\usepackage{SCITEPRESS}     

\usepackage[outdir=./]{epstopdf}
\usepackage[colorinlistoftodos,prependcaption,textsize=tiny]{todonotes}
\newcommand{\matr}[1]{\mathbf{#1}}     

\begin{document}

\title{Automatic Estimation of Sphere Centers from Images of Calibrated Cameras}

\author{\authorname{Levente Hajder \sup{1}
, Tekla T\'oth \sup{1}
and Zolt\'an Pusztai \sup{1}
}
\affiliation{\sup{1}Department of Algorithms and Their Applications, E\"otv\"os Lor\'and University,\\ P\'azm\'any P\'eter stny. 1/C, Budapest, Hungary, H-1117}
\email{\{hajder,tekla.toth,puzsaii\}@inf.elte.hu}
}




\keywords{Ellipse Detection, Spatial Estimation, Calibrated Camera, 3D Computer Vision}

\abstract{
  Calibration of devices with different modalities is a key problem in robotic vision. Regular spatial objects, such as planes,  are frequently used for this task. This paper deals with the automatic detection of ellipses in camera images, as well as to estimate the 3D position of the spheres corresponding to the detected 2D ellipses. We propose two novel methods to (i) detect an ellipse in camera images and (ii) estimate the spatial location of the corresponding sphere if its size is known. The algorithms are tested both quantitatively and qualitatively. They are applied for calibrating the sensor system of autonomous cars equipped with digital cameras, depth sensors and LiDAR devices.    
}

\onecolumn \maketitle \normalsize \setcounter{footnote}{0} \vfill

\section{\uppercase{Introduction}}
\label{sec:introduction}

\noindent Ellipse fitting in images has been a long researched problem in computer vision for many decades~\cite{Proffitt1982}. Ellipses can be used for camera calibration~\cite{Ji2001CameraSF,879788}, estimating the position of parts in an assembly system~\cite{DetectionRHTAuto} or for defect detection in printed circuit boards (PCB)~\cite{CLu}. This paper deals with a very special application of ellipse fitting: detected ellipses are applied for calibration of multi-sensor systems via estimating the spatial locations of the spheres corresponding to the ellipses.

Particularly, this paper concentrates on two separate problems: (i) automatic and accurate ellipse detection is addressed first, then (ii) the spatial location of the corresponding sphere is calculated if the radius of the sphere is known.

\noindent \textbf{Ellipse detection.} There are several solutions for ellipse fitting, but only a few of those can detect the ellipse contours accurately as well as robustly. Traditional algorithms can be divided into two main groups: (i) \textit{Hough transform} and  (ii) \textit{Edge following}.

Hough transform (HT) based methods for ellipse detection tend to be slow. A general ellipse has five degrees of freedom and it is found by an exhaustive search on the edge points. Each edge pixel in the image votes for the corresponding ellipses~\cite{Duda1972}. Therefore, evaluating the edge pixel in the five-dimensional parameter space has high computation and memory costs. Probabilistic Hough Transform (PHT) is a variant of the classical HT: it randomly selects a small subset of
the edge points which is used as input for HT~\cite{Kiryati1991}.
The 5D parameter space can be divided into two pieces. First, the ellipse center is estimated, then the remaining three parameters are found in the second stage~\cite{Yuen1989,Tsuji1978}.

Edge following methods try to connect the line segments, usually obtained by the widely-used Canny edge detector~\cite{Canny1986,Plataniotis2000}. These segments are refined in order to fit to the curve of an ellipse. The method of Kim et
al.~\cite{Kim2002} merges the short line segments to longer arc segments,
where the arc fitting algorithms are frequently called. Mai et al. published~\cite{Chia2011} another method based on similar idea, the difference lies in linking the segments and edge points by adjacency and curvature conditions.

Lu et al.\ detect images based-on arc-support lines~\cite{CLu}. First, arc-support groups are formed from line segments, detected by the Canny~\cite{Canny1986} or Sobel detectors~\cite{kanopoulos1988design}. Then an initial ellipse set generation and ellipse clustering is applied to remove the duplicated ones. Finally, a candidate verification process removes some of the candidates and re-fit the remaining ones.

\cite{Basca2005RandomizedHT} proposed the Randomized Hough Transform (RHT) for ellipse detection. Their work is based on~\cite{HTEllipse}, but achieve significantly faster detection time. In addition to randomization, further filtering methods are applied to remove false detections.

The method introduced by Fornaciari et al.\ approaches real-time performance~\cite{FornaciariFastAndEffective}. The method detects the arc from Sobel derivatives and classifies them according their convexity. Based on their convexity,  mutual positions and implied ellipse center, the arcs are grouped and the ellipse parameters are estimated. Finally, parameters clustering is applied to duplicated ellipses.
%

Nowadays, deep-learning methods~\cite{Jin2019} are very popular, they can also be applied for ellipse detection. However, the authors of this paper thinks that ellipse detection is a pure geometric problem, therefore deep-learning solutions are not the best choice for this task. The problem is not only the lack of sufficiently large training data, but also the time demand at training. Thus, deep-learning based methods are considered as an overkill for ellipse detection.


\noindent \textbf{Sphere location in 3D.} The main application area of the proposed ellipse detector is the calibration of different sensors, especially range sensors and cameras. Usually, chessboards~\cite{Geiger2012} or other planar calibration targets~\cite{Park} are applied for this task, however, recently spherical calibration objects~\cite{Kummerle2018} has also begun to be used for this purpose. The basic idea is that a sphere can be accurately and automatically detected on both depth sensors and camera images. Extrinsic parameters can than computed by point-set registration methods~\cite{Arun1987} if at least four sphere centers are localized. Unfortunately, detection of planar targets~\cite{Geiger2012} are inaccurate due to the sparsity of the point cloud, measured by a depth camera or LiDAR device.

The theoretical background of our solution is as follows: (i) an ellipse determines a cone in 3D space; (ii) if the radius of this sphere is known, the 3D location of the sphere can be computed using the fact that the cone is tangent to the sphere.

Our method differs from that of Kummerle et al.~\cite{Kummerle2018} in the sense that 3D sphere and 2D ellipse parameters are analytically determined. The 3D position of the sphere is directly computed from these parameters contrary to~\cite{Kummerle2018} in which the 3D estimation is based on the perspective projection of the reconstructed sphere and the edges are used to tune the parameters.

\noindent \textbf{Contribution.} 
The novelty of the paper is twofold: (i) A novel, robust ellipse estimation pipeline is proposed that yields accurate ellipse parameters. It does not have parameters to be set, it is fully automatic. (ii) A 3D location estimation procedure is also proposed. It consists of two steps: first, rough estimation for the sphere location is given, then the obtained coordinates are refined via numerical optimization. Both steps and substeps are novel to the best of our knowledge.

\noindent \textbf{Structure of the paper.} Section~\ref{sec:theoretical} shows how a sphere is perspectively projected onto an image, and how the ellipse parameters, corresponding to the contour points, are obtained. Section~\ref{sec:ell_det} introduces the RANSAC threshold for ellipse estimation. Then, Section~\ref{sec:proposed_methods} describes our method for ellipse detection and 3D center estimations. Section~\ref{sec:experiments} evaluates the performance of the proposed method against state-of-the-art techniques, and finally, Section~\ref{sec:conclusion} concludes our work.

\section{Theoretical Background: Perspective Projection of Sphere onto Camera Image}
\label{sec:theoretical}

\noindent In this section, we show how the spatial location of the sphere determines the ellipse parameters in the images. 

A calibrated pin-hole camera is assumed to be used, thus the intrinsic parameters are known. Let $\mathbf{K}$ denote the camera matrix. Its elements are as follows:

\begin{equation}
\mathbf{K}=\left[\begin{array}{ccc}
f_{u} & 0 & u_{0}\\
0 & f_{v} & v_{0}\\
0 & 0 & 1
\end{array}\right],
\end{equation}
where $f_u$, $f_v$, and location $[u_0\quad v_0]^T$ are the horizontal and vertical focal length and the principal point~\cite{Hartley2003}, respectively.

Without loss of generality, the coordinate system is assumed to be aligned to the camera. 
%
%
The location of a pixel in the image is denoted by the 2D vector $[u,v]^{T}$. 
Then the normalized image coordinates are calculated as follows: 
\begin{equation}
\left[\begin{array}{c}
\hat{u}\\
\hat{v}\\
1
\end{array}\right]=\matr K^{-1}\left[\begin{array}{c}
u\\
v\\
1
\end{array}\right]=\left[\begin{array}{c}
\frac{u-u_{0}}{f_{u}}\\
\frac{v-v_{0}}{f_{v}}\\
1
\end{array}\right],
\end{equation}
where $[\hat{u}, \hat{v}]^T$ denotes the normalized coordinate of $[{u}, {v}]^T$.

The back-projection of a pixel corresponds to a ray in the spatial space. The spatial points of the ray can be written as $\left[x, y, z\right]^T=z\left[\hat{u},\hat{v},1\right]^T$.
%
%
Using the implicit equation of the sphere, i.e. $(x-x_0)^2+(y-y_0)^2+(z-z_0)^2=r^2$,  the intersection of a ray and the sphere can be determined as
\begin{equation}
\left(z\hat{u}-x_{0}\right)^{2}+
\left(z\hat{v}-y_{0}\right)^{2}+
\left(z-z_{0}\right)^{2}
=r^{2},
\end{equation}
where $r$ and $[x_0,y_0,z_0]^T$ denote the radius and the center of the sphere, respectively. It is straightforward that the previous equation is a quadratic function in $z$:
\begin{gather}
    \left(\hat{u}^2+\hat{v}^2+1\right)z^2- 
    2\left(\hat{u}x_{0}+\hat{v}y_{0}+z_{0}\right)z+ \nonumber \\
    x_{0}^{2}+y_{0}^{2}+z_{0}^{2}-r^{2}
    =0.
\end{gather}
The quadratic function has a single root at the border of the sphere when the ray is tangent to the spherical surface. Thus, the discriminant of the quadratic equation equals to zero in this case:
\begin{gather}
    \left(-2\left(\hat{u}x_{0}+\hat{v}y_{0}+z_{0}\right)\right)^2- \nonumber \\
    4\left(\hat{u}^2+\hat{v}^2+1\right)\left( x_{0}^{2}+y_{0}^{2}+z_{0}^{2}-r^{2}\right)
    =0.
\end{gather}
This constraint to the determinant can be rearranged to the implicit form of a conic section $A\hat{u}^{2}+B\hat{u}\hat{v}+C\hat{v}^{2}+D\hat{u}+E\hat{v}+F=0$, where the coefficients are as follows:
%
\begin{flalign} 
\label{coeff}
A &=  r^{2}-y_{0}^{2}-z_{0}^{2}, & B &= 2x_{0}y_{0}, & C &=  r^{2}-x_{0}^{2}-z_{0}^{2},  \nonumber \\ 
 D &=  2x_{0}z_{0}, & E &=  2y_{0}z_{0},  & F &= r^{2}-x_{0}^{2}-y_{0}^{2}. 
\end{flalign}
%

%
%
%
\begin{figure}[t]
    \centering
    \includegraphics[width=.98\linewidth]{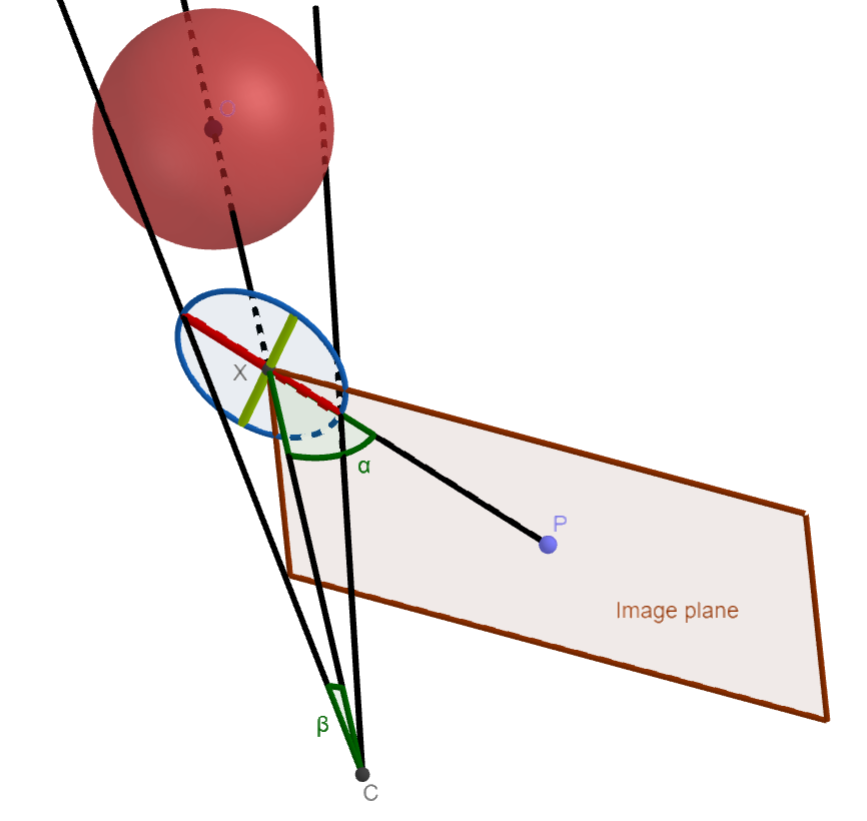}
    \caption{The 3D scene of the sphere projection into the image plane in case of the analyzed far-out situation.}
    \label{fig:3d_sphere_proj}
\end{figure}
\section{\uppercase{Ellipse approximation with a circle}}
\label{sec:ell_det}
\noindent 
One critical part of the sphere center estimation is to detect the projection of the sphere in the images. As we proved in Sec.~\ref{sec:theoretical}, this shape is a conic section, usually an ellipse. One particular case is visualized in Fig~\ref{fig:3d_sphere_proj}, where the intersection point of the ellipse axes lands one of the corners of the image. To robustly fit an ellipse to image points, either HT or RANSAC~\cite{RANSAC} has to be applied, but both methods are slow. The former needs to accumulate results of the 5D parameter space, and the latter needs a large iteration number. It principally depends on the inlier/outlier ratio and the number of parameters to be estimated. A general ellipse detector needs $5$ points, however, only $3$ points are needed for circle fitting. Therefore, the iteration number of RANSAC can be significantly reduced by estimating the ellipse with a circle at the first place.

This section introduces a RANSAC threshold selection process for circle fitting.
The threshold should be larger than the largest distance between the ellipse and circle contours but this threshold has to be small enough not to classify to many points 
as false positive inliers. 
The first condition is fulfilled by our definition of the threshold. Theoretically obtained values in Fig.~\ref{fig:real_input_ransac_thr} show that our approach which based on the described method in Sec.~\ref{sec:threshold_circle} satisfies the second condition as well.

Therefore, one of the novelty in this paper is to show that circle fitting with higher threshold can be applied for ellipse fitting. The next section introduces how this threshold can be selected.

\subsection{Threshold Selection for Circle Fitting}
\label{sec:threshold_circle}


Circle model is a rough approximation for ellipses in robust RANSAC-based~\cite{RANSAC} fitting. Although the error threshold for the inlier selection has a paramount importance, it is difficult do define it on a manual way. Basically, this threshold is needed because of the noise in datapoints. The main idea here is that ellipses can be considered as noisy circles. Realistic camera setups confirm this assumption because if the spherical target is not too close to the image plane, then the ratio of the ellipse axes will be close to one. This is convenient in spite of the most extreme visible position: when the ellipse is as far as possible from the principal point: near one of the image corners as it is visualized in Fig.~\ref{fig:3d_sphere_proj}. 
 
We propose an adaptive solution to find threshold $t$ based on the ratio $s$ of the minor and major axes, denoted by $a$ and $b$. In our algorithm, the coherence of the variables is defined as $s=\frac{R+t}{R-t}$, where $R$ is the radius of the actually fitted circle in the image. Hence, the searched variable is $t=\frac{(s-1)R}{s+1}$. Because of realistic simulation, we have to add the noise in pixels, too.

\begin{figure}
    \centering
    \includegraphics[width=.98\linewidth]{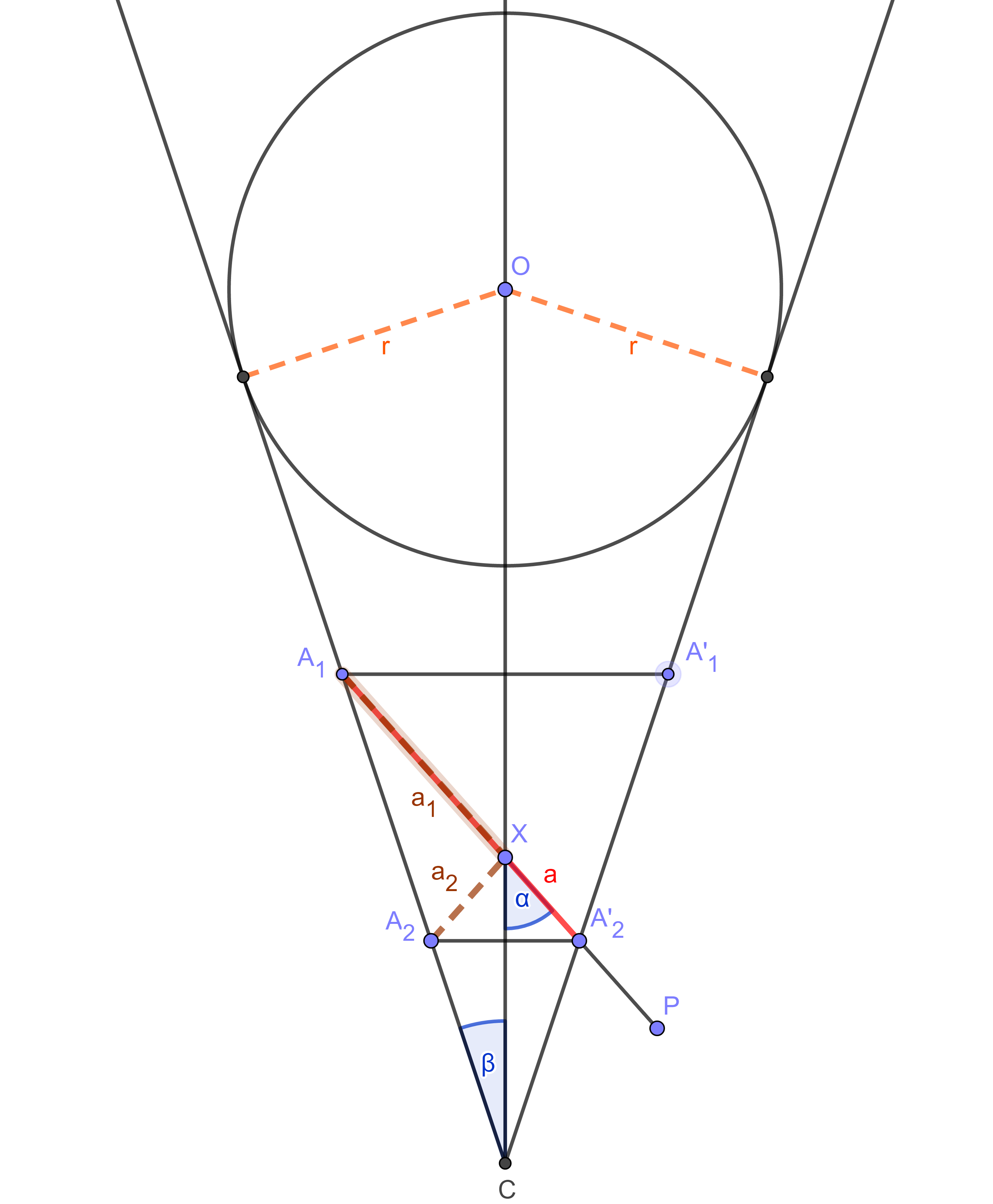}
    \caption{Planar segment of the cone which contains the major axis of the ellipse.}
    \label{fig:2d_major}
\end{figure}

\begin{figure}
    \centering
    \includegraphics[width=.98\linewidth]{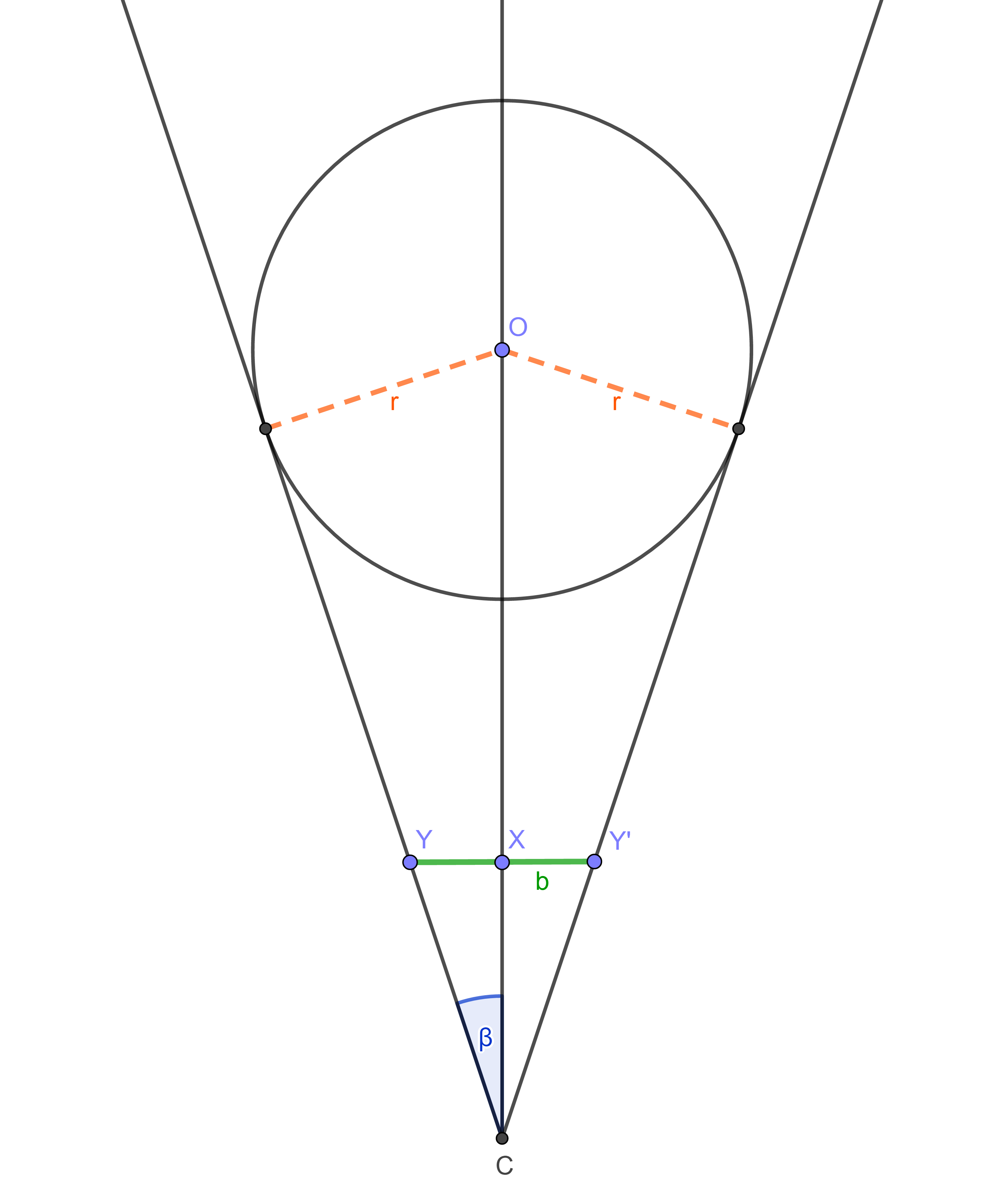}
    \caption{Planar segment of the cone which contains the minor axis of the ellipse.}
    \label{fig:2d_minor}
\end{figure}

\begin{figure}
    \centering
    \includegraphics[width=.98\linewidth]{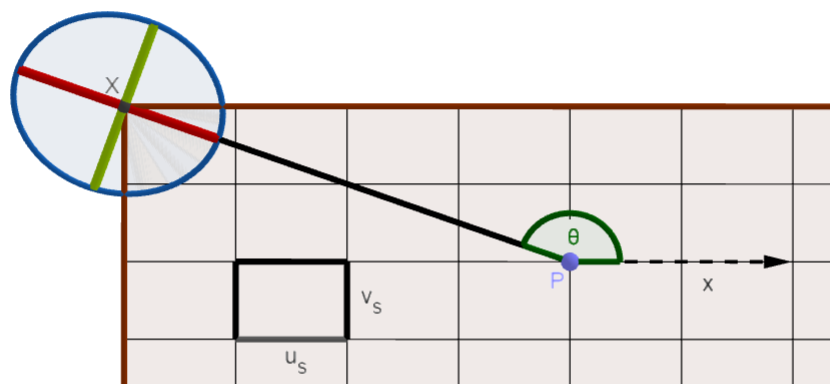}
    \caption{If the ratio of the pixel width and height is $u_s : v_s$, where $u_s \neq v_s$, the ratio $\frac{a}{b}$ has to be scaled. This modification depends on the pixel scaling factors $u_s$, $v_s$ and the angle $\theta$ between the axis $x$ and the main axis of the ellipse.}
    \label{fig:pix_scale}
\end{figure}


The discussed 3D problem is visualized in Fig. \ref{fig:3d_sphere_proj} where not the whole ellipse is fall into the area of the image. Because of the main goal is to estimate $s$, two angles has to be defined: angle $\alpha$ between the axis of the cone and the image plane and angle $\beta$ between a generator of the cone and the axis of the cone. 
To find this two angles, consider the two plane segments of the cone which contain the axis of the cone and either the minor or the major axis of the ellipse. The 2D visualization is pictured in Fig.~\ref{fig:2d_major} and~\ref{fig:2d_minor}. This two views are perpendicular to each other because the axes of the ellipse are perpendicular. Fig.~\ref{fig:2d_minor} shows that the minor axis and the axis of the cone are also perpendicular. 
If the ellipse center gets further from the principal point then $\alpha$ is smaller, and the difference between $a$ and $b$ steadily increases as it is seen in Fig.~\ref{fig:2d_major}. However, if $\alpha = \frac{\pi}{2}$, the major axis is also perpendicular to the axis of the cone: this is the special case, when $a=b$ and the projection is a circle which center is the principal point. 

First, the vectors and distances have to be defined which determinate $\alpha$ and $\beta$. The 2D coordinates of the intersection point $\mathbf{X}$ of the ellipse axes in the image is denoted by $\begin{bmatrix}u & v\end{bmatrix}^T$ , the camera position is $\mathbf{C}=\begin{bmatrix}0 & 0 & 0\end{bmatrix}^T$ and the distance of the image plane to the camera in the camera coordinate system is $Z=1$. The 3D coordinates of $\mathbf{X}$ become
\begin{eqnarray}
\mathbf{X}=Z \mathbf K^{-1}\left[\begin{array}{c}
u\\
v\\
1
\end{array}\right]=
Z \left[\begin{array}{c}
\frac{u-u_{0}}{f_{u}}\\
\frac{v-v_{0}}{f_{v}}\\
1
\end{array}\right].
\end{eqnarray}
The angle $\alpha$ between the axis of the cone and the image plane is determined by two vectors: the direction vector from $\mathbf{X}$ to $\mathbf{C}$, and the vector from $\mathbf{X}$ to the principal point 
$\mathbf{P}=f \mathbf K^{-1}\begin{bmatrix}u_{0}\\
v_{0}\\
1
\end{bmatrix}=f\begin{bmatrix}0\\
0\\
1
\end{bmatrix}$, because it is known that the line of the main axis of the conic section contains the principal point of the image. The vector coordinates become

\begin{equation}
    \mathbf{v=\mathbf{X\mathbf{C}}=}Z \left[\begin{array}{c}
\frac{u_{0}-u}{f_{u}}\\
\frac{v_{0}-v}{f_{v}}\\
-1
\end{array}\right] , \quad 
\mathbf{p}=\mathbf{X}\mathbf{P}=f \left[\begin{array}{c}
\frac{u_{0}-u}{f_{u}}\\
\frac{v_{0}-v}{f_{v}}\\
0
\end{array}\right].
\end{equation}
%
Using vector $\mathbf{v}$ and $\mathbf{p}$, angle $\alpha$ between image plane and cone axis is calculated as
\begin{equation*}
\alpha=acos\left(-\left[\frac{\mathbf{v}}{\begin{Vmatrix}\mathbf{v}\end{Vmatrix}}\right]\cdot\left[\frac{\mathbf{p}}{\begin{Vmatrix}\mathbf{p}\end{Vmatrix}}\right]\right).
\end{equation*}
Similarly, the angle between the generator and the axis of the cone depends on the distance of the sphere center to the camera $d=\begin{Vmatrix}\mathbf{O}-\mathbf{C}\end{Vmatrix}$ and the radius of the sphere $r$,  shown in Fig. ~\ref{fig:2d_major} or ~\ref{fig:2d_minor}. In our approach, $r$ is given, however, $d$ is undefined because of the varying position of the sphere. The distance is estimated using the focal length, the radius of the sphere and the radius of the fitted ellipse: $d = \frac{r f}{R}$. Considering the two scalars $d$ and $r$, the angle becomes:
\begin{equation}
\beta=\arcsin\left(\frac{r}{d}\right).
\end{equation}

\begin{figure*}[t]
    \centering
    \includegraphics[width=.31\linewidth]{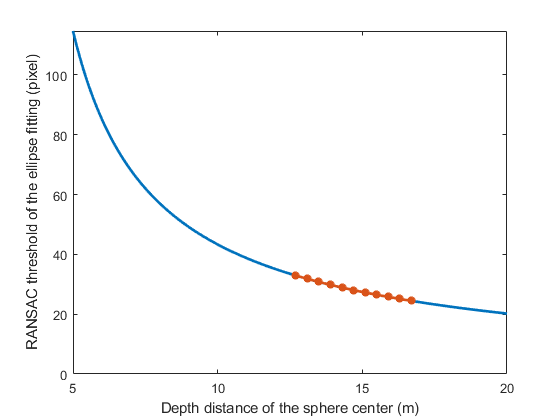}
    \includegraphics[width=.31\linewidth]{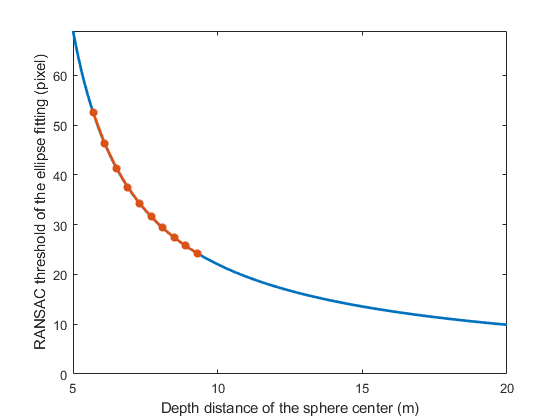}
    \includegraphics[width=.31\linewidth]{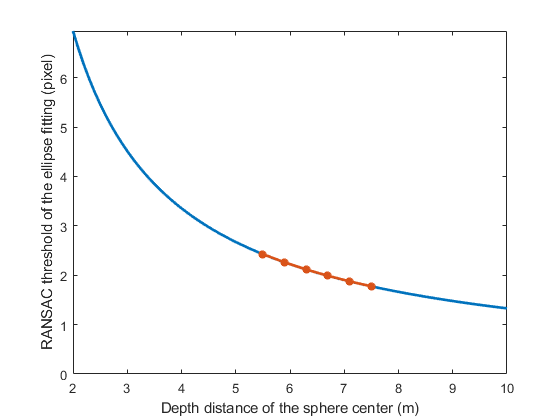}
    \includegraphics[width=.31\linewidth]{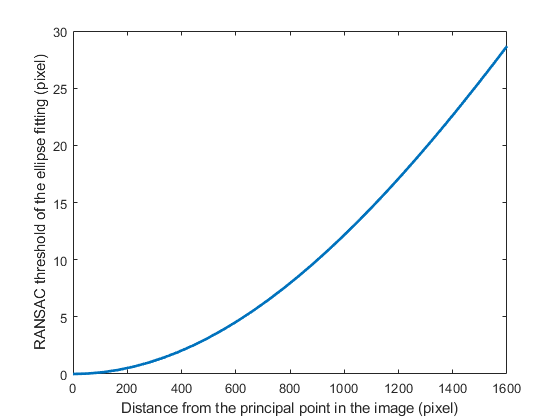}
    \includegraphics[width=.31\linewidth]{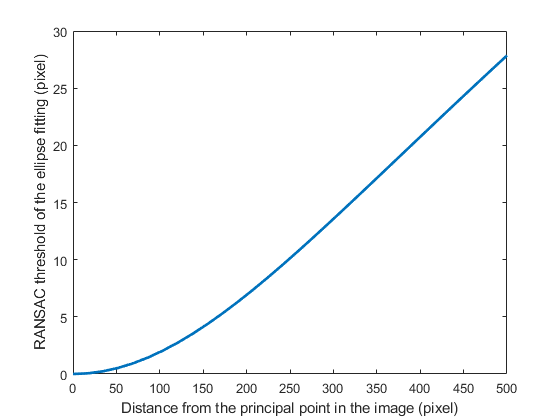}
    \includegraphics[width=.31\linewidth]{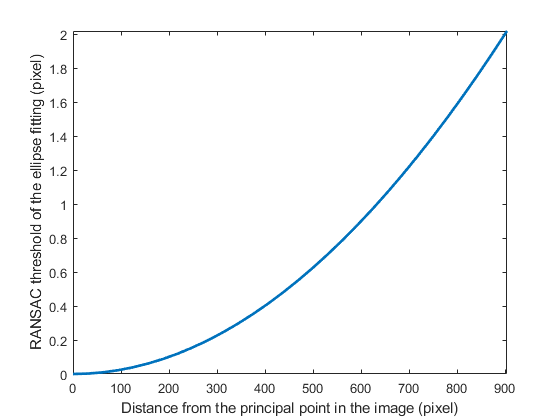}
        \caption{Estimated RANSAC threshold with varying depth (top) and varying distance from the principal point in the image plane (down) in case of Blensor tests (left), Carla tests (middle) and real word tests (right). Red colored part of the curves in the first row denotes the range of the applied values in the tests based on the measured depth of the spheres.}
    \label{fig:real_input_ransac_thr}
\end{figure*}

The searched ratio can be estimated using the angles $\alpha$ and $\beta$ and the special properties of the projections as it is discussed in App. ~\ref{sec:app:ratio} :
\begin{gather}
    \frac{a}{b}= 
    \frac{s_\alpha c^{2}_\beta}{s^{2}_\alpha c^{2}_\beta-c^{2}_\alpha s^{2}_\beta}.
\end{gather}
However, the pixels can be scaled horizontally and vertically as it is visualized in Fig. ~\ref{fig:pix_scale}. The ratio $s$ depends on this scaling factors ${u_{s}}$, ${v_{s}}$ and the angle $\theta$ between the direction vector $\mathbf{x} = \begin{bmatrix}1 & 0 & 0\end{bmatrix}^T$ along axis $x$ and the main axis $a$ of the ellipse:
\begin{equation}
\theta=acos\left(-\left[\frac{\mathbf{v}}{\begin{Vmatrix}\mathbf{v}\end{Vmatrix}}\right]\cdot\mathbf{x}\right).
\end{equation}
After applying elementary trigonometrical expressions, the modified scale becomes
\begin{gather}
  s=
  \frac{a\sqrt{u_{s}^{2}c^{2}_\theta+v_{s}^{2}s^{2}_\theta}}{b\sqrt{u_{s}^{2}c^{2} \left( \frac{\pi}{2}+\theta \right) +v_{s}^{2}s^{2} \left( \frac{\pi}{2}+\theta \right) }}= \nonumber \\
  \frac{a}{b}\sqrt{\frac{u_{s}^{2}c^{2}_\theta+v_{s}^{2}s^{2}_\theta}{u_{s}^{2}s^{2}_\theta+v_{s}^{2}c^{2}_\theta}}= \nonumber \\
  \frac{s_\alpha c^{2}_\beta}{s^{2}_\alpha c^{2}_\beta-c^{2}_\alpha s^{2}_\beta}\sqrt{\frac{u_{s}^{2}c^{2}_\theta+v_{s}^{2}s^{2}_\theta}{u_{s}^{2}s^{2}_\theta+v_{s}^{2}c{2}_\theta}}  .
\end{gather}
Then the RANSAC threshold $t=\frac{(s-1)R}{s+1}$ can be computed using the detected circle radius $R$ and the calculated ratio $s$.

Fig.~\ref{fig:real_input_ransac_thr} shows the validation of the automatic RANSAC threshold estimation with varying depth of the sphere center and with varying distance between the ellipse center and the principal point in the image plane in three different test environments, which are detailed in Sec.~\ref{sec:experiments}. The red curve segments show the range of the applied thresholds in our tests. The thresholds are realistic and successfully applied in our ellipse detection method.

\section{\uppercase{Proposed algorithms}}
\label{sec:proposed_methods}

\noindent In this section, the proposed algorithms for spatial sphere estimation is overviewed. Basically, the estimation consists of two parts: the first one finds the contours of the ellipse in the image, the second part determines the 3D location of the sphere corresponding to the ellipse parameters. Simply speaking, the 2D image processing task has to be solved first, then 3D estimation of the location is achieved.

\subsection{Ellipse Detector}

The main challenge for ellipse estimation is that a real image may contain many contour lines that are independent of the observed sphere. There are several techniques to find an ellipse in the images as it is overviewed in the introduction. 

Our ellipse fitting method is divided into several subparts:
\begin{enumerate}
    \item Edge points are detected in the images.
    \item The largest circle is found in the processed image by RANSAC-based~\cite{RANSAC} circle fitting on the edge points.
    \item Then the edges with high magnitude around the selected circle is collected.
    \item Finally, another RANSAC~\cite{RANSAC}-cycle is run in order to robustly estimate the ellipse parameters. 
\end{enumerate}

\noindent \textbf{Edge point detection.} RANSAC-based algorithms~\cite{RANSAC} usually work on data points. For this purpose, 2D points from edge maps are retrieved. The Sobel operator~\cite{Plataniotis2000} is applied for the original image, then the strong edge points, i.e. points with higher edge magnitude than a given threshold, are selected. The edge selection is based on edge magnitudes, however, edge directions are also important as the tangent directions of the contours of ellipses highly depend on the ellipse parameters. Finally, the points are made sparser: if there are more strong points in a given window, only one of those are kept.

\noindent \textbf{Circle fitting.} This is the most critical substep of the whole algorithm. As it is proven in Sec.~\ref{sec:threshold_circle}, the length of the two principal axes of the ellipse is close to each other. As a consequence, a circle-fitting method can be applied, and the threshold for RANSAC, overview in Sec.~\ref{sec:threshold_circle} has to be applied. 

\begin{figure}
    \centering
    \includegraphics[width=.49\linewidth]{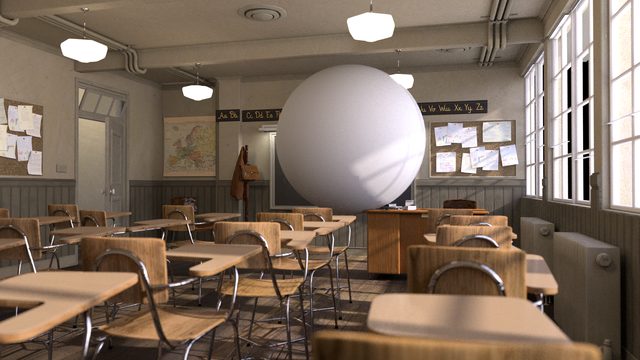}
    \includegraphics[width=.49\linewidth]{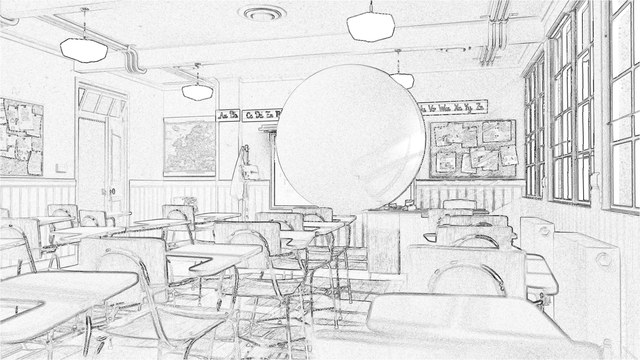}

    \caption{Left: Original image generated by Blender. Right: Edge magnituded after applying the Sobel  operator.}
    \label{fig:example_blender}
\end{figure}

The main challenge is that only minority of the points are related to the circle. As it is pictured in the left image of Fig.~\ref{fig:example_blender}, only $5-10 \%$ of the points can be considered as inliers for circle fitting. Therefore, a very high repetition number is required for RANSAC. In our practice, millions of iterations are set to obtain acceptable results for real or realistic test cases.

An important novelty of our circle detector is that the edge directions are also considered: edge directions of inliers must be orthogonal to the tangent of the circle curve.

\begin{figure}
    \centering
    \includegraphics[width=.99\linewidth]{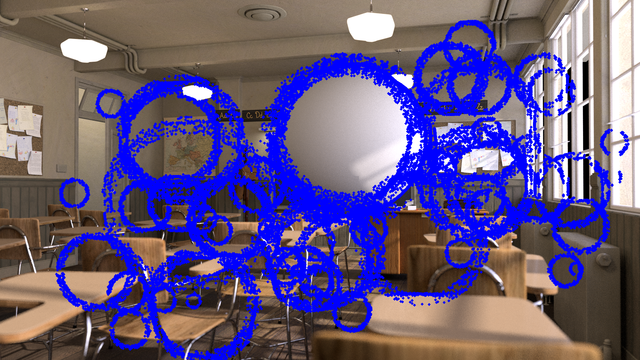}

    \caption{Candidate points for ellipse fitting. After edge-based circle fitting, the strongest edge points are selected radially for the candidate circles.}
    \label{fig:points_for_fitting}
\end{figure}

\begin{figure}
    \centering
    \includegraphics[width=.49\linewidth]{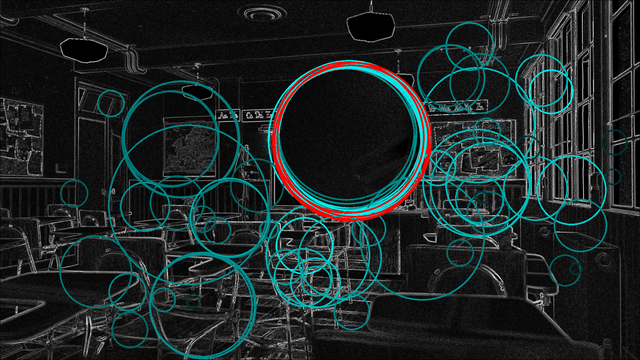}
    \includegraphics[width=.49\linewidth]{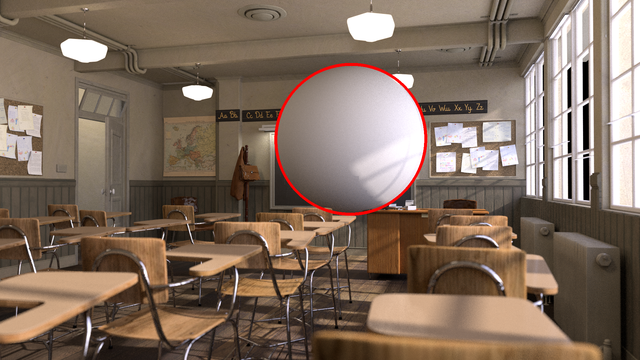}

    \caption{Left: Candidate ellipses. Red color denotes the circle with the highest score. Right: Final ellipse.}
    \label{fig:points_for_fitting_final}
\end{figure}

\noindent \textbf{Collection of strong edge points.}
The initial circle estimation yields only a preliminary estimation. For the final estimation, the points are relocated. For this purpose, the proposed algorithm searches the strongest edge points along radial direction. An example for the candidate points are visualized in the right image of Fig.~\ref{fig:points_for_fitting}. 

\noindent \textbf{Final ellipse fitting.}
The majority of the obtained points belong to the ellipse. However, there can be incorrect points, effected by e.g. shadows or occlusion. Therefore, robustification of the final ellipse fitting is a must. Standard RANSAC~\cite{RANSAC}-based ellipse fitting is applied in our approach. We select the method of Fitzgibbon et al.~\cite{Fitzgibbon1999} in order to estimate a general ellipse.
A candidate ellipse is obtained then for each circle. Figure~\ref{fig:points_for_fitting_final} shows the candidate ellipses on the left. There are many ellipses around the correct solutions, the similar ellipses have to be found: the final parameters are selected as the average of those. An example for the final ellipse is visualized in the right picture of Fig.~\ref{fig:points_for_fitting_final}.

\subsection{3D Estimation}
When the ellipse parameters are estimated, the 3D location of the sphere can be estimated as well if the radius of the sphere is known. All points and the corresponding tangent lines of the ellipse in conjunction with the camera focal points determine tangent planes of the ellipse.
If there is an ellipse, represented by quadratic equation
\begin{equation*}
    Ax^2+Bxy+Cy^2+Dx+Ey+F=0,
\end{equation*}
it can be rewritten into matrix form if the points are written in homogeneous coordinates as
\begin{equation*}
\left[\begin{array}{ccc}
x & y & 1\end{array}\right]\left[\begin{array}{ccc}
A & \frac{B}{2} & \frac{D}{2}\\
\frac{B}{2} & C & \frac{E}{2}\\
\frac{D}{2} & \frac{E}{2} & F
\end{array}\right]\left[\begin{array}{c}
x\\
y\\
1
\end{array}\right]=0
\end{equation*}
The inner matrix, denoted by $\matr T$ in this paper, determines the ellipse. The tangent line $\matr l$ at location $\left[ x\quad y \right]^T$ can be determined by $\matr l=\matr T \left[\begin{array}{ccc}
x & y & 1\end{array}\right]^T$ as it is discussed in~\cite{Hartley2003}. Then the tangent plane of the sphere, corresponding to this location, can be straightforwardly determined.

\begin{figure}
    \centering
    \includegraphics[width=.98\linewidth]{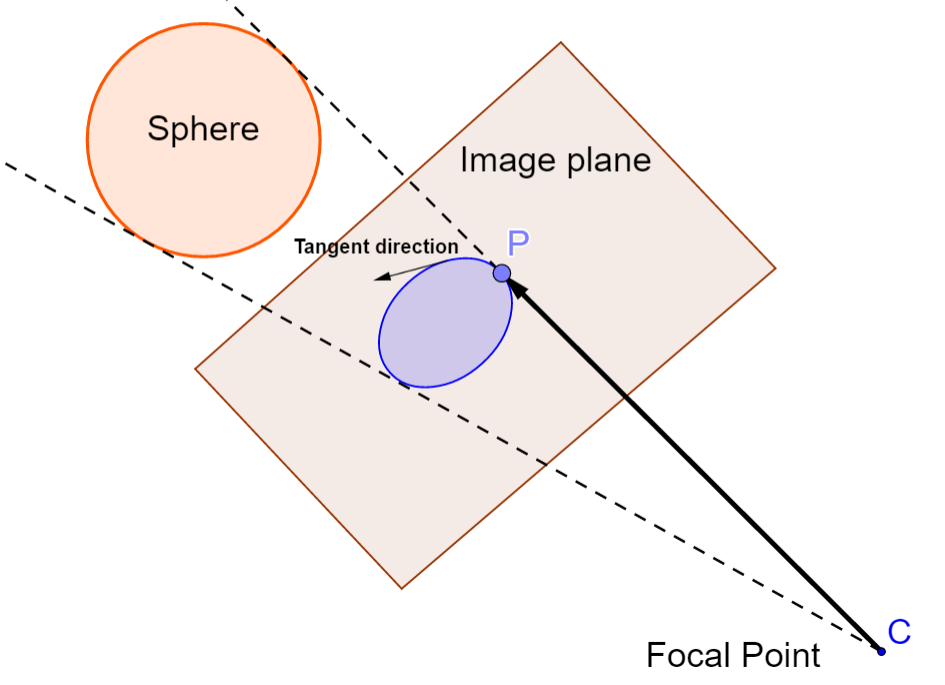}
    \caption{The 3D center estimation problem. }
    \label{fig:problem}
\end{figure}

The 3D problem is visualized in Fig.~\ref{fig:problem}. The tangent plane of the sphere is determined by the focal point $C$. Two of its  tangent vectors are represented by the tangent direction $\mathbf l$ of the ellipse in the image space~\footnote{Third coordinate is zero in 3D.}, and the vector $\overline{CP}$. The plane normal is the cross product of the two tangent vectors.

If a point and the normal, denoted by $\matr p_i$ and $\matr n_i$, respectively, of the tangent plane are given, the distance of the sphere center with respect to this plane is the radius $r$ itself. If the length of the normal is unit, in other words $\matr n_i^T \matr n_i = 1$, the distance can be written as
\begin{equation*}
    r= \matr n_i^T \left( \matr p_i - \matr x_0 \right),
\end{equation*}
where $\matr x_0$ is the center of the sphere, that is the unknown vector of the problem. Each tangent plane gives one equation for the center. If there are three tangent planes, the location can be determined. In case of at least four planes, the problem is over-determined. The estimation is given via an inhomogeneous linear system of equations as follows:
\begin{equation}
    \left[\begin{array}{c}
\matr n_{1}^{T}\\
\matr n_{2}^{T}\\
\vdots\\
\matr n_{N}^{T}
\end{array}\right] \matr x_{0}=\left[\begin{array}{c}
\matr n_{1}^{T} \matr p_{1}-r\\
\matr n_{2}^{T} \matr p_{2}-r\\
\vdots\\
\matr n_{N}^{T} \matr p_{N}-r
\end{array}\right]
\label{eq:3D_est}
\end{equation}
For the over-determined case, the pseudo-inverse of the matrix has to be computed and multiplied with the vector on the right side as the problem is an inhomogeneous linear one~\cite{Bjorck1996}.

\noindent \textbf{Numerical optimization.} As the 3D estimation of the sphere center, defined in Eq.~\ref{eq:3D_est}, minimizes the distances in 3D between sphere center and tangent planes, it is not optimal since the detection error appears in the image space. Therefore, we apply finally numerical optimization for estimating the sphere center. Initial values for the final optimization are given by the geometric minimization, then the final solution is obtained by a numerical optimization. Levenberg-Marquardt technique is used in our approach.

\section{\uppercase{Experiments}}
\label{sec:experiments}

\noindent We have tested the proposed ellipse fitting algorithm both on three different testing scenarios:
\begin{itemize}
    \item \textit{Synthesized test.} In order to test the accuracy and numerical stability of the algorithms, we have tested the proposed fitting and 3D estimation algorithms on a synthetic environment. We have selected Octave for this purpose. As the 3D models are generated by the testing environment, quantitative evaluation can be carried out. The weak-point of the full synthetic test is that only point coordinates are generated, therefore our image-processing algorithm cannot be tested.
    \item \textit{Semi-synthetic test.} Semi-synthetic tests extend the full-synthetic one by images. There are rendering tools, applied usually for Computer Graphics (CG) applications, that can generate images of a know virtual 3D scene. We have used Blender\footnote{http://blender.org} as it is one of the best open-source CG tools and can produce photo-realistic images. As the virtual 3D models are known, ground truth sphere locations as well as camera parameters can be retrieved from the tool, therefore quantitative evaluation of the algorithms is possible.
    
    The calibration of different devices is very important for autonomous system, therefore we have tried the proposed methods for an autonomous simulator. We selected an open-source simulator, called CARLA~\footnote{www.carla.org}, for this purpose. It is based on the widely-used UnReal Engine.
    
    \item \textit{Real test.} Even the most accurate CG tools cannot generate full-realistic images, therefore real testing cannot be neglected. 
    The GT data in these test are obtained by using a novel sphere fitting algorithm, designed for LiDAR data processing~\cite{visapp19} ,and by LiDAR-camera calibration. 
\end{itemize}

\subsection{Synthetic tests}

Synthetic test was only constructed in order to validate that the formula, given in Eq.~\ref{eq:3D_est}, is correct. For this purpose, a simple synthetic testing environment was implemented in Octave~\footnote{Octave is an open-source MATLAB-clone. See http://www.octave.org.}. Camera parameters as well as the ground-truth (GT) sphere parameters were randomly generated, ellipse parameter was computed by projecting the points into the camera image as it is overviewed in Section~\ref{sec:theoretical}. 

\noindent \textbf{Conclusion of synthetic test.} It was successfully validated that Equation~\ref{eq:3D_est} is correct, the GT sphere parameters can always be exactly retrieved. 

\subsection{Semi-synthetic tests}
\begin{figure*}
    \centering
    \includegraphics[width=.49\columnwidth]{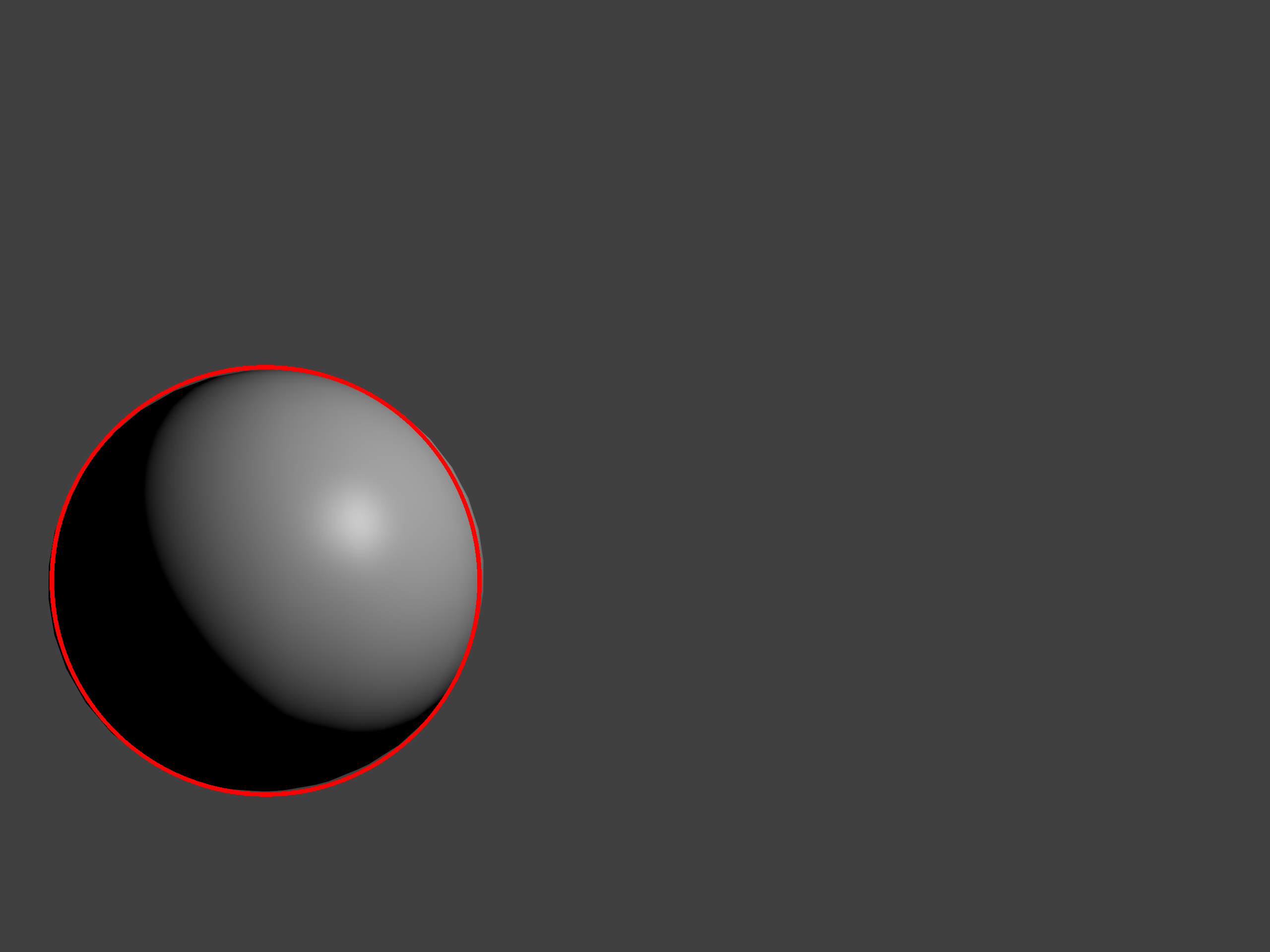}
    \includegraphics[width=.49\columnwidth]{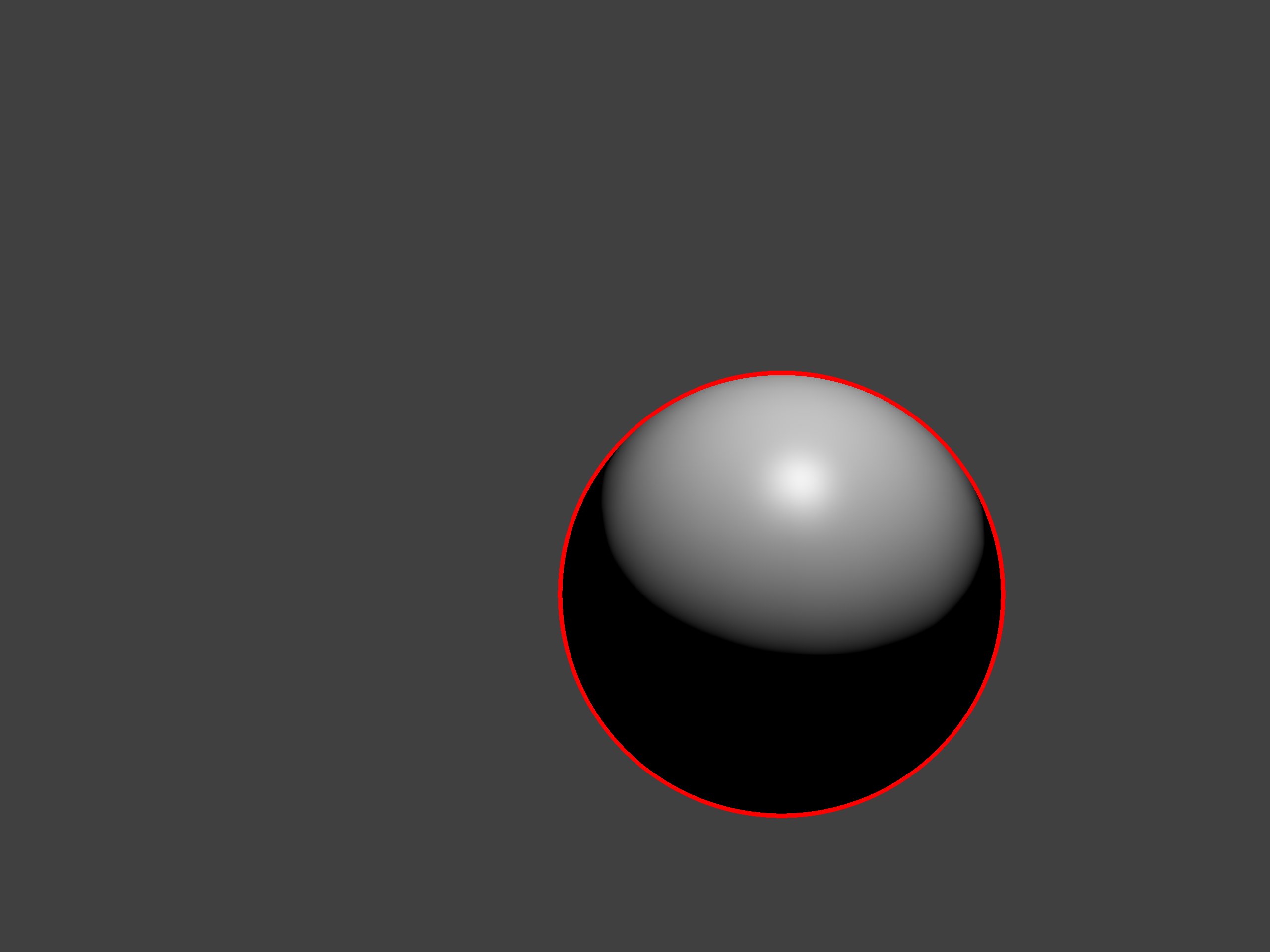}
    \includegraphics[width=.49\columnwidth]{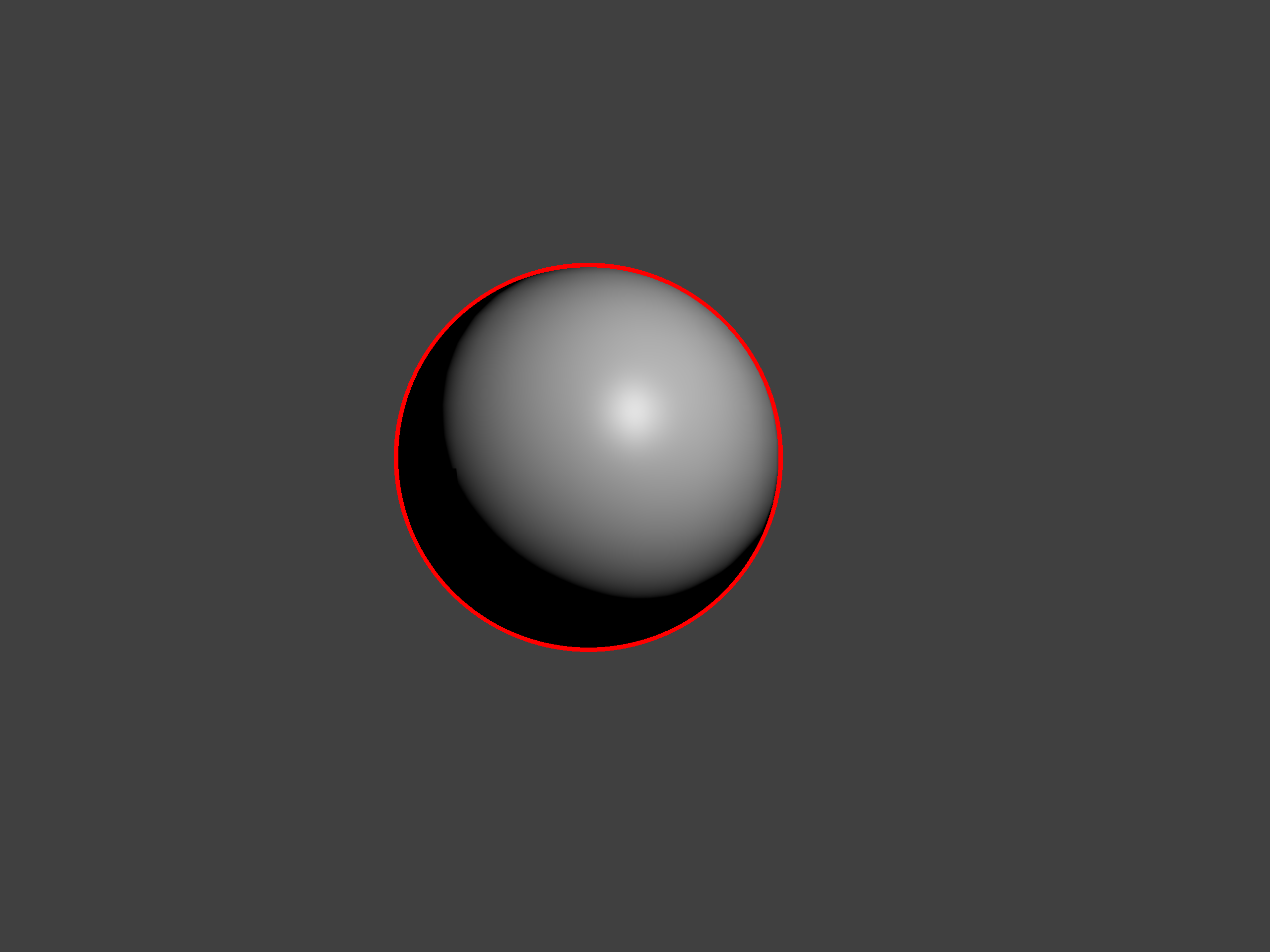}
    \includegraphics[width=.49\columnwidth]{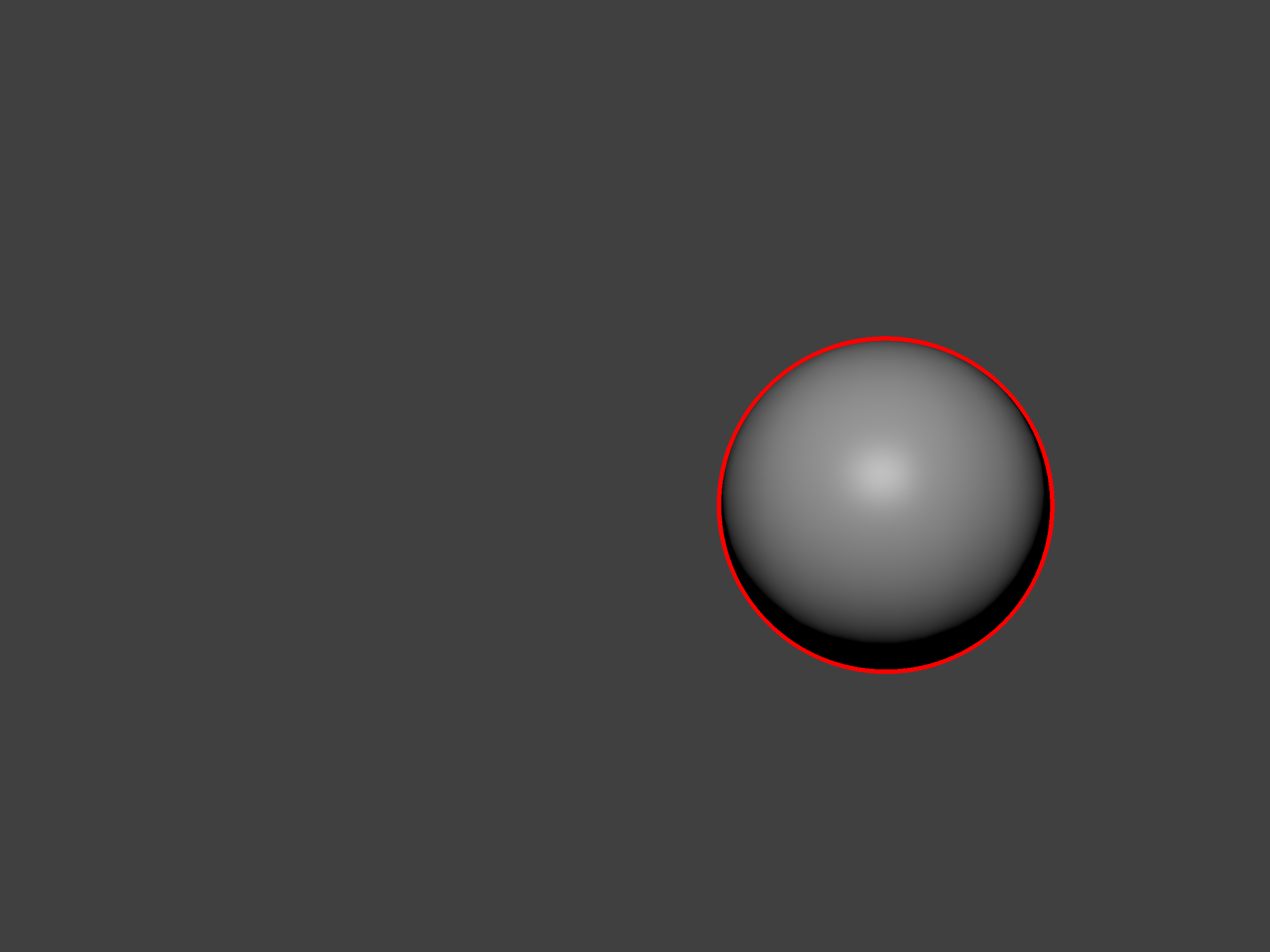}    
    \caption{Simple test sequences rendered by Blender. Detected ellipses visualized by red.}
    \label{fig:Blender_simple}
\end{figure*}

\begin{figure*}
    \centering
    \includegraphics[width=.49\columnwidth]{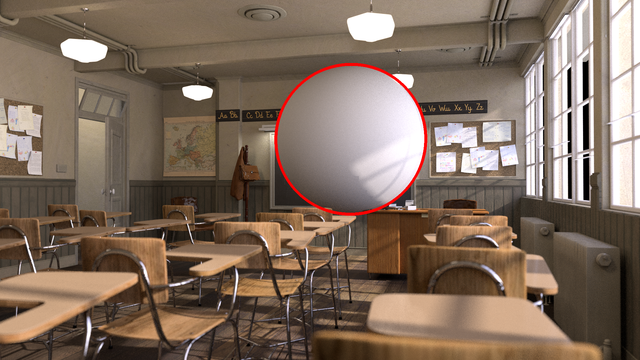}
    \includegraphics[width=.49\columnwidth]{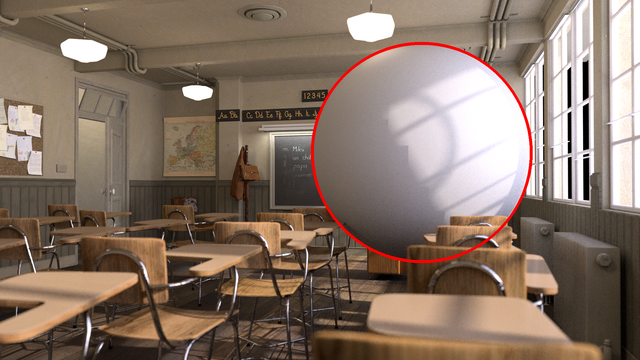}
    \includegraphics[width=.49\columnwidth]{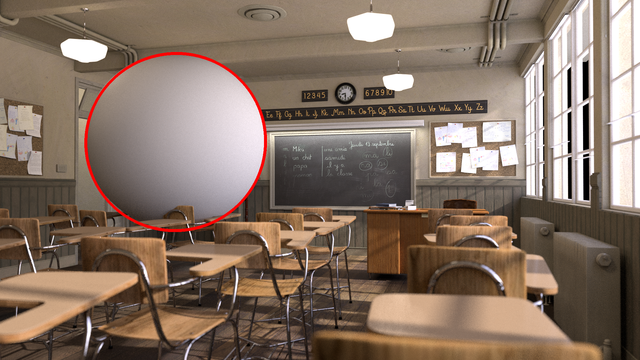}
    \includegraphics[width=.49\columnwidth]{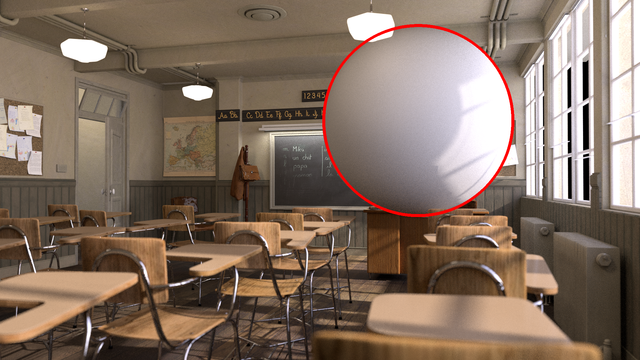}
    \caption{Complex 'classroom' test sequences rendered by Blender. Detected ellipses visualized by red.}
    \label{fig:Blender_classroom}
\end{figure*}

The semi-synthetic tests include virtual scenes containing simple shaded spheres. An example is visualized in the left image of Fig.~\ref{fig:example_blender}. The images are generated by the well-known free and open source 3D creation suite Blender\footnote{http://www.blender.org}. We have generated four test images. They are visualized in Fig.~\ref{fig:Blender_simple} and Fig.~\ref{fig:Blender_classroom}.

Two scenes are generated by Blender. The first contains only a single sphere with Phong-shading only. It is visualized in Fig.~\ref{fig:Blender_simple}. This is considered as an easier test case, because the images contain only a single ellipse. The second scene is the well-known classroom scene, which is widely used in computer graphics papers. The scene contains several objects and a sphere. Therefore, the synthetic images are rich in edges which makes the detectors work harder. The images are visualized in Fig.~\ref{fig:Blender_classroom}.

\begin{figure*}
    \centering
    \includegraphics[width=.49\columnwidth]{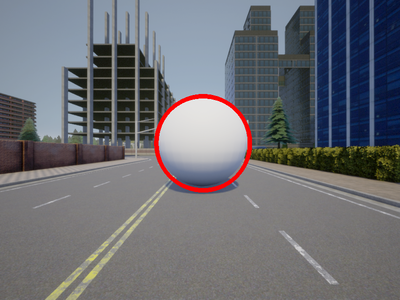}
    \includegraphics[width=.49\columnwidth]{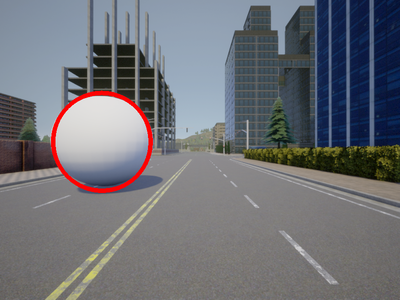}
    \includegraphics[width=.49\columnwidth]{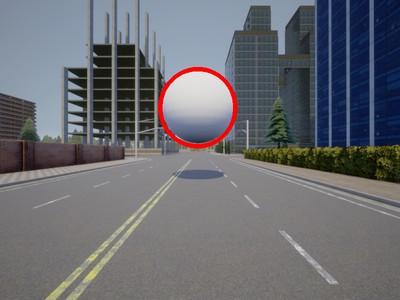}
    \includegraphics[width=.49\columnwidth]{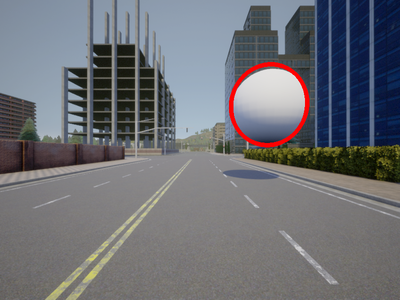}
\caption{Test images generated by autonomous simulator called Carla. The quality of the spheres are not high due to the rough tessellation of the sphere. Detected ellipses visualized by red. }
    \label{fig:Carla}
\end{figure*}

Since the application of these algorithm for LiDAR-camera calibration is also important, one additional scene is rendered by the CARLA simulator. This scene contains a typical driving simulation with an additional sphere on the road. It is visualized in Fig.~\ref{fig:Carla}.

\subsection{Real tests}

\begin{figure*}
    \centering
    \includegraphics[width=.19\linewidth]{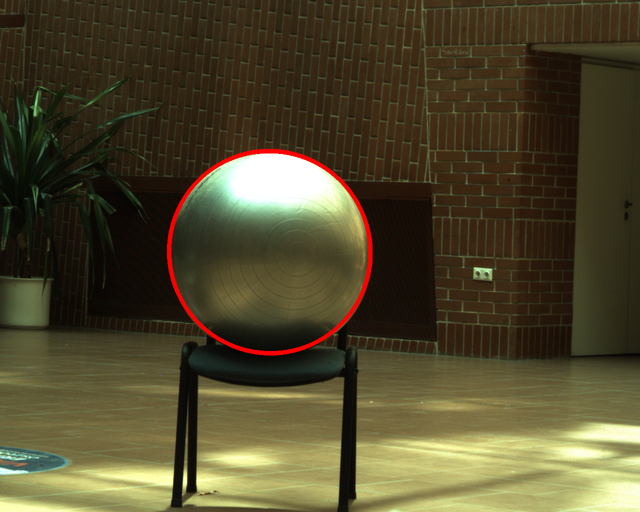}
    \includegraphics[width=.19\linewidth]{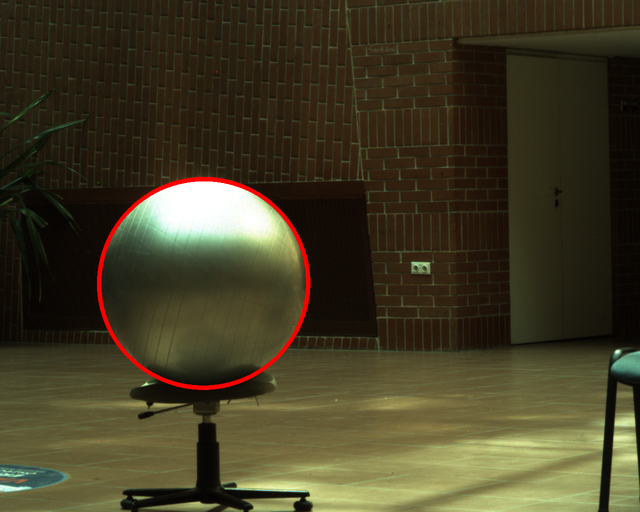}
    \includegraphics[width=.19\linewidth]{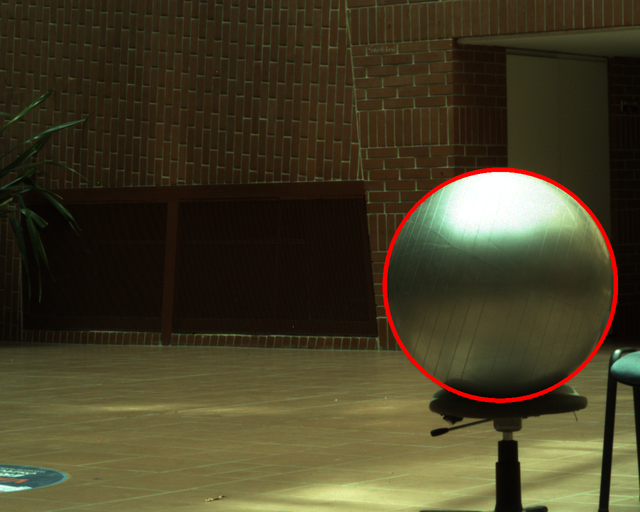}
    \includegraphics[width=.19\linewidth]{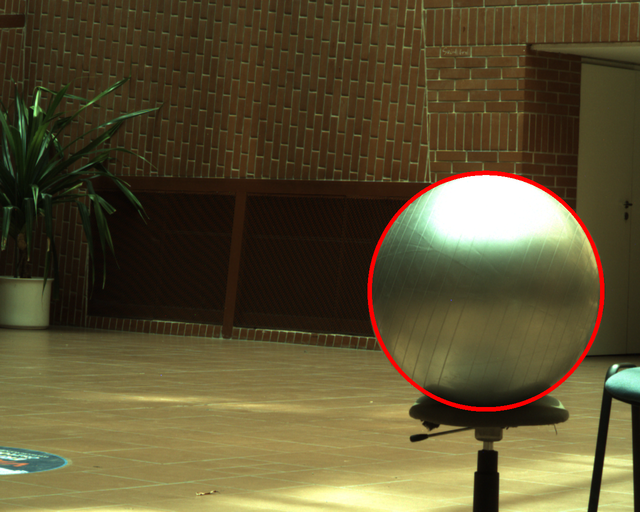}
    \includegraphics[width=.19\linewidth]{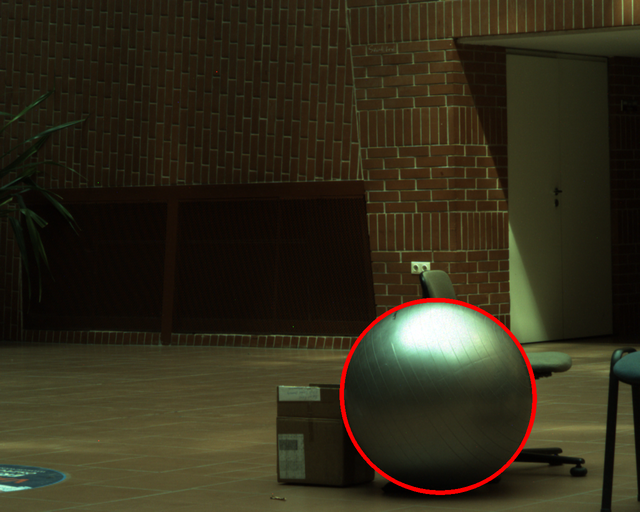}
        \caption{Input images for real tests.}
    \label{fig:real_input}
\end{figure*}

The main challenge for a realistic test is to find a good large sphere. We have selected a gymnastic plastic ball for this purpose, with a radius of $30 cm$. The images were taken by an iCube camera, whose intrinsic parameters were calibrated using the well known Zhang-calibration technique~\cite{Zhang2000}. Then radial distortion was removed, and the algorithms were run with camera parameters $f_u=f_v=4529$ (focal lenght; optics have narrow field of view), $u_0=659$, $v_0=619$. (principal point; camera sensor resolution is $1280 \times 1024$). Fig.~\ref{fig:real_input} shows the images, and the detected ellipses.

\subsection{Evaluation}

The proposed ellipse detector is compared to four state-of-the-art (SoA) methods. The error of the methods measured as the Euclidean distance between the GT spatial sphere centers and the estimated ones calculated from the ellipse parameters. 

In the real-world tests, the GT is obtained by fitting a sphere to the LiDAR data using a novel method~\cite{visapp19}, that is tailored for LiDAR datapoints. Finally, the sphere centers are transformed from the LiDAR coordinate system to the camera coordinate system. The GT data in the synthetic tests are obtained from the testing environment.

Four SoA are compared to the proposed method. These are as follows:
\begin{enumerate}
    \item \textit{FastAndEffective}: \cite{FornaciariFastAndEffective} proposed a method that can be used for real-time ellipse detection. First, arch groups are formed from the Sobel derivatives, then the ellipse parameters are estimated. The main problem with this method is the large number of parameters, which have to be tuned for each image sequence individually.
    \item \textit{Random Hough Transform (RHT)}: The method introduced by~\cite{Basca2005RandomizedHT} reduced the computation time of the HT by randomization. The method achieves lower detection accuracy, since it considered only the edge point positions, not their gradient.
    \item \textit{HighQuality}: \cite{CLu} proposed this method, which results high quality ellipse parameters. They generate arcs from Sobel or Canny edge detectors, and several properties of these arcs are exploited, e.g.\ overall gradient distribution, arc-support directions or polarity. This methods need $3$ parameters to be tuned for each test sequence.
    \item \textit{FastInvariant}: \cite{Jia16} trade off accuracy for further speed improvements. Their method removes straight arcs based on line characteristic and obtains ellipses by conics.
\end{enumerate}

Table~\ref{tab:testresults} shows the results of both the synthetic and real-world tests. The rows containing the integer values are the different image indices in the same test environment, and the Euclidean error is shown for every method. In some of the cases, the methods were not able to find the right ellipses in the images, even after careful parameter tuning. In these cases, the error of the method is not presented. The accuracy of the proposed method (denoted by the rows beginning with Prop.), is comparable to the SoA. The best methods is clearly the \textit{HighQuality} in all test cases, however, it was not able to find any ellipse in the 5-th image of the real-world test. While \textit{RHS} achieves the worst accuracy, the \textit{FastInvariant} and \textit{FastAndEffective} methods have almost the same results. \textit{RHS} needs to know the approximated size of the major axis and the ratio between the minor and major axis of the ellipse in pixels. The latter two methods require more than eight parameters to be set. Even though the proposed method does not achieve significantly better results then the others, it is the only completely parameter-free, thus fully automatic, method. 

%
%

\begin{table*}
    \centering
    \begin{tabular}{c|c c c c||c|c c c c c}
            Blensor & 1 & 2 & 3 & 4 & Real & 1 & 2 & 3 & 4 & 5 \\
             \hline
            Prop. & 0.6463 & \textbf{0.0100} & 0.0372 & 0.2327 & Prop. & 0.9115 & 1.1317 & 1.7169 & 0.8702 &\textbf{ 0.4443} \\
            FAE & 0.0660 & 0.1087 & 0.0601 & 1.7892 & FAE & 0.6926 & \textbf{0.2386} & \textbf{0.1566} & \textbf{0.1618} & -\\ 
            RHS &  0.7831 & 0.0537 &  0.0396 & 0.1907 & RHS & 0.6449 & 0.3330 & 0.3781 & 0.1667 & 0.7859\\
            HQ & \textbf{0.0516} &  0.0527 & \textbf{0.0135} & \textbf{0.0263} & HQ & \textbf{0.5231} & 0.4587 & 0.3239 & 0.1734 & -\\ 
            FI & 0.0660 & 0.1087 & 0.0601 & 1.7892 & FI & 0.6926 & \textbf{0.2386} & \textbf{0.1566} & \textbf{0.1618} & -\\
            \hline
            \hline
    \end{tabular}
    \begin{tabular}{c|c c c c c c c c}
            Classroom & 1 & 2 & 3 & 4 & 5 & 6 & 7 & 8 \\
            \hline
            Prop. &  0.016 & 0.0213 & 0.0302 & - & 0.2094 & - & 0.0555 & -\\
            FAE & 0.0168 & 0.0168 & 0.0682 & 0.0553 & - & 0.0985 & 0.1650 & - \\
            RHS & 0.0689 & 0.0689 & 0.1698 & 0.1165 & 0.6551 & 0.41658 & 7.8553 & 0.0574 \\
            HQ & \textbf{0.0141} & \textbf{0.0141} & \textbf{0.0037} & \textbf{0.0058} & \textbf{0.0571} & \textbf{0.0056} & \textbf{0.03422} & \textbf{0.0081} \\
            FI & 0.0168 & 0.0168 & - & 0.0553 & - & 0.0985 & 0.1650 & - \\
            \hline
            \hline
            Carla & 1 & 2 & 3 & 4 & 5 & 6 & 7 & 8 \\
            \hline
            Prop. & 0.2993 & - & 1.5381 & 2.6420 & 0.3870 & 1.4974 & 3.1761 & 0.55548 \\
            FAE & - & 0.1846 & 0.1038 & 2.2196 & 0.1074 & 0.6769 & 0.5710 & 0.2690\\
            RHS & 0.2260 & 2.17628 & 1.8085 & 0.8335 & 1.0599 & 0.3954 & \textbf{0.0733} & 0.32469 \\
            HQ & \textbf{0.0586} & \textbf{0.0486} & \textbf{0.0434} & \textbf{ 0.0740} & \textbf{0.0933} & \textbf{0.0836 }& 0.1260 & \textbf{0.0391 }\\
            FI & - & 0.1846 & 0.1038 &2.2196 &0.1074 &0.4107 &0.5710 & - 
    \end{tabular}
    \caption{Test results for the synthetic and real-world tests. The top rows with the numbers mark the test cases for each test environments (Blensor, Classroom, Carla and Real), The results are measured Euclidean error between the GT and estimated ellipse centers by the applied methods. (The notations are : Prop. = Proposed method, FAE = FastAndEffective, RHS = Random Hough Transform,  HQ = HighQuality and FI = FastInvariant.)}
    \label{tab:testresults}
\end{table*}

\section{\uppercase{Conclusions}}
\label{sec:conclusion}
This paper proposes a novel 3D location estimation pipeline for spheres. It consists of two steps: (i) the ellipse is estimated first, (ii) then the spatial location is computed from the ellipse parameters, if the radius of the sphere is given and the cameras are calibrated. Our ellipse detector is accurate as it is validated by the test. The main benefit of our approach is that it is fully automatic as all parameters, including the RANSAC~\cite{RANSAC} threshold for circle fitting, are fixed in the implementation. To the best of our knowledge, our second method, i.e. the estimator for surface location, is a real novelty in 3D vision. The main application area of our pipeline is to calibrate digital cameras to LiDAR devices and depth sensors.

\vfill
\section*{\uppercase{Acknowledgements}}
\noindent T. Tóth and Z. Pusztai were supported by the project \texttt{EFOP-3.6.3-VEKOP-16-
2017-00001}: Talent Management in Autonomous Vehicle Control Technologies, by the Hungarian Government and co-financed by the European
Social Fund.
L. Hajder was supported by the project no. \texttt{ED\_18-1-2019-0030} (Application domain specific highly reliable IT solutions subprogramme). It has been implemented with the support provided from the National Research, Development and Innovation Fund of Hungary, financed under the Thematic Excellence Programme funding scheme.

\bibliographystyle{apalike}
{\small
\bibliography{bib/SfM,bib/Hajder,bib/normal,bib/ellipse,bib/LidarCalib}}

\clearpage
\appendix

\section*{\uppercase{Appendix}}

\noindent \section{Ellipse fitting to 2D points.}

The task is to estimate an ellipse if $N$ data points are given. The ellipse is given by its implicit form
\begin{equation*}
Ax^2+Bxy+Cy^2+Dx+Ey+F=0,
\end{equation*}
where parameters $A$ -- $F$ represent the quadratic curve. If there are $N$ datapoints: $\left( [x_1\quad y_1]^T,\dots, [x_N\quad y_N]^T \right)$, the algebraic fitting problem can be written as $\arg \min_{\mathbf G} \left| \mathbf G \mathbf x \right|^2_2$, where
\begin{equation*}
\mathbf G=\left[\begin{array}{cccccc}
x_{1}^{2} & x_{1}y_{1} & y_{1}^{2} & x_{1} & y_{1} & 1\\
\vdots & \vdots & \vdots & \vdots & \vdots & \vdots\\
x_{N}^{2} & x_{N}y_{N} & y_{N}^{2} & x_{N} & y_{N} & 1
\end{array}\right],
\mathbf x=\left[\begin{array}{c}
A\\
B\\
C\\
D\\
E\\
F
\end{array}\right]
\end{equation*}

The scale of the parameters are arbitrary selected. Quadratic curves can be ellipses, parabolas or hiperbolas. For the former case, the constraint $B^2-4AC < 0$ has to be hold. This can be written by the quadratic equation $\matr x^T \matr H \matr x= -1$, where

\begin{equation*}
\matr H = \left[\begin{array}{cccccc}
0 & 0 & -2 & 0 & 0 & 0\\
0 & 1 & 0 & 0 & 0 & 0\\
-2 & 0 & 0 & 0 & 0 & 0\\
0 & 0 & 0 & 0 & 0 & 0\\
0 & 0 & 0 & 0 & 0 & 0\\
0 & 0 & 0 & 0 & 0 & 0
\end{array}\right] .
\end{equation*}
Then the scale ambiguity of the parameters is also eliminated since $B^2-4AC =-1$
If a Lagrangian multiplier $\lambda$ is introduced, the constrained problem, i.e. the algebraic fitting of an ellipse itself, can be solved by the generalized eigenvalue technique~\cite{Fitzgibbon1999} as $\matr J x=\lambda \matr H x$, where $\matr J= \matr G^T \matr G$.

Remark that the normalization of data points are usually important in order to get accurate results. There are two viable solutions: (i) data normalization is applied as it is discussed in App.~\ref{app:norm}, or (ii) the method if Halir \& Flusser~\cite{Halir1998} is used. In our tests, we selected the latter solution.

\section{Data normalization for ellipse fitting}
\label{app:norm}
As it is frequently appeared for numerical methods, data normalization is critical for efficient parameter estimation. If the coordinates in coefficient matrix $\matr G$ come from pixel row/column number, then the order of magnitude for different rows are significantly different in the column of $\matr G$. For example, the first, second, and last columns contain elements around $10^6$, $10^3$, and $10^0$, respectively. 

A usual technique for data normalization is to move the origin to the center of the gravity of the points, represented by the vector $[u_0,v_0]^T$ , and then scale the data. These transformations can be represented by the product of $3 \times 3$ matrices as
\begin{eqnarray*}
\matr T_N=\left[\begin{array}{ccc}
s_{1} & 0 & 0\\
0 & s_{2} & 0\\
0 & 0 & 1
\end{array}\right]& \left[\begin{array}{ccc}
1 & 0 & u_{0}\\
0 & 1 & v_{0}\\
0 & 0 & 1
\end{array}\right]=\\ \left[\begin{array}{ccc}
s_{1} & 0 & s_{1}u_{0}\\
0 & s_{2} & s_{2}v_{0}\\
0 & 0 & 1
\end{array}\right] = &   \left[\begin{array}{ccc}
\hat f_u & 0 & \hat u_{0}\\
0 & \hat f_v  & \hat v_{0}\\
0 & 0 & 1
\end{array}\right].
\end{eqnarray*}

If the original estimation ellipse is written as
\begin{equation*}
Ax^{2}+Bxy+Cy^{2}+Dx+Ey+F=0
\end{equation*}
then the normalized one is as follows:
\begin{eqnarray*}
A'\left(\hat f_{u}x+\hat u_{0}\right)^{2}+B'\left(\hat f_{u}x+\hat u_{0}\right)\left(\hat f_{v}y+\hat v_{0}\right)+C'\left(\hat f_{v}y+\hat v_{0}\right)^{2}+\\D'\left(\hat f_{u}x+
\hat u_{0}\right)+E'\left(\hat f_{v}y+\hat v_{0}\right)+F'=0.
\end{eqnarray*}

The conversion between the original and normalized ellipse parameters are as follows:
\begin{eqnarray*}
A=&A' \hat f_{u}^{2}\\
B=&B' \hat f_{u}f_{v}\\
C=&C' \hat f_{v}^{2}\\
D=&2A' \hat f_{u} \hat u_{0}+B' \hat f_{u} \hat v_{0}+D' \hat f_{u}\\
E=&B' \hat f_{v} \hat u_{0}+2C' \hat f_{v} \hat v_{0}+E' \hat f_{v}\\
F=&A' \hat u_{0}^{2}+B' \hat u_{0} \hat v_{0}+C' \hat v_{0}^{2}+D' \hat u_{0}+E' \hat v_{0}+F'
\end{eqnarray*}

Vice versa:

\begin{eqnarray*}
A'=&\frac{A}{\hat f_{u}^{2}}\\
B'=&\frac{B}{\hat f_{u} \hat f_{v}}\\
C'=&\frac{C}{\hat f_{v}^{2}}\\
D'=&\frac{D-2A' \hat f_{u} \hat u_{0}-B' \hat f_{u} \hat v_{0}}{\hat f_{u}}\\
E'=&\frac{E-2C' \hat f_{v} \hat v_{0}-B' \hat f_{v} \hat u_{0}}{ \hat f_{v}}\\
F'=&F-A' \hat u_{0}^{2}-B' \hat u_{0} \hat v_{0}-C' \hat v_{0}^{2}-D' \hat u_{0}-E' \hat v_{0}
\end{eqnarray*}

\section{Ratio of ellipse axes}
\label{sec:app:ratio}

\begin{figure}[t]
    \centering
    \includegraphics[width=.98\linewidth]{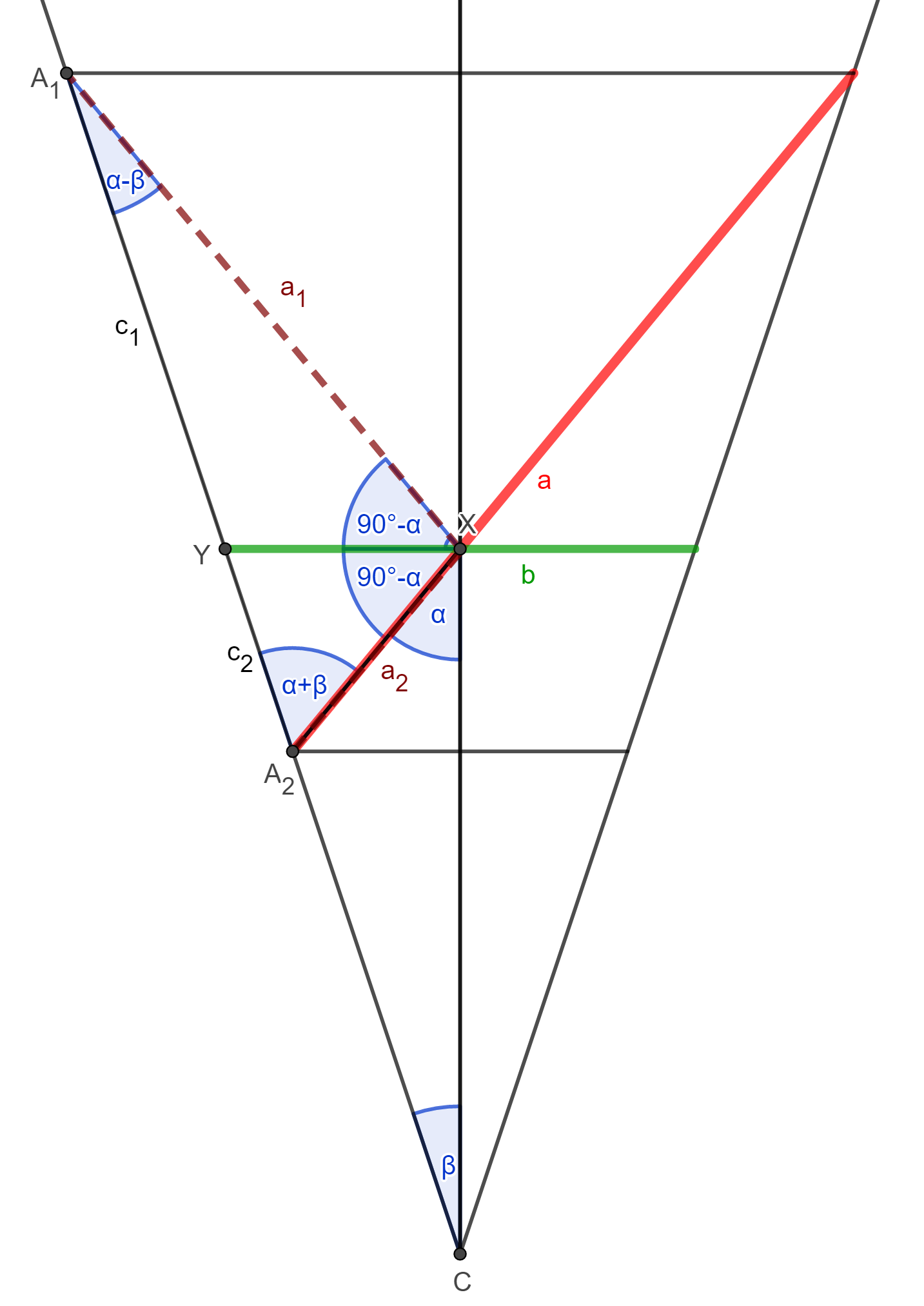}
    \caption{The plane segments of Figure ~\ref{fig:2d_major} and ~\ref{fig:2d_minor} projected into one image containing both the minor and major axes of the ellipse. Using the known angles $\alpha$ and $\beta$ the searched ratio $\frac{a}{b}$ can be computed.}
    \label{fig:2d_sphere_proj}
\end{figure}

The ratio of the axes of an ellipse determinates the RANSAC threshold of circle fitting in our method as it is discussed in Section~\ref{sec:ell_det}. To find this ratio $\frac{a}{b}$, we project the two conic sections from Fig. ~\ref{fig:2d_major} and ~\ref{fig:2d_minor} into one figure rotating one of the sections with $\frac{\pi}{2}$. The result of the projection is visualized in Fig. ~\ref{fig:2d_sphere_proj}. The length of the perpendicular section to the main axis of the cone along $\mathbf{X}$ will be $b$ in every view, hence the combined figure, which contains both $a$ and $b$ supplemented by known angles looks like in this image.

If $a_{1}$ and $a_{2}$ denote the two parts of the section $a$, containing  $\mathbf X$, the axes of the ellipse and their ratio depend on $\alpha$ and $\beta$ as $a=a_{1}+a_{2}$ and $c=c_{1}+c_{2}$.

Because of the law of cosines in triangles $A_{1}XY$ and $A_{2}XY$, and applying the angle bisector rule in triangle $A_{1}A_{2}X$, the following formula can be written:
\begin{equation}
\label{eq:c1}
c_{1}^{2}=a_{1}^{2}+\frac{b^{2}}{4}-2a_{1}\frac{b}{2}\cos \left( \frac{\pi}{2}-\alpha \right),
\end{equation}
\begin{equation}
\label{eq:c2}
c_{2}^{2}=a_{2}^{2}+\frac{b^{2}}{4}-2a_{2}\frac{b}{2}\cos \left( \frac{\pi}{2}-\alpha \right),   
\end{equation}
\begin{equation}
\label{eq:c1c2}
\frac{a_{1}}{c_{1}}=\frac{a_{2}}{c_{2}}.    
\end{equation}
Substituting Eqs.~\ref{eq:c1} and~\ref{eq:c2} into Eq.~\ref{eq:c1c2} yields:
\begin{equation}
    \frac{a_{1}^{2}}{a_{2}^{2}}=\frac{a_{1}^{2}+\frac{b^{2}}{4}-2a_{1}\frac{b}{2}cos \left( \frac{\pi}{2}-\alpha \right) }{a_{2}^{2}+\frac{b^{2}}{4}-2a_{2}\frac{b}{2}\cos \left( \frac{\pi}{2}-\alpha \right)}.
\end{equation}
As $b\neq0$ it can be reordered:
\begin{gather}
   a_{1}^{2}\left(a_{2}^{2}+\frac{b^{2}}{4}-2a_{2}\frac{b}{2} \cos \left( \frac{\pi}{2} -\alpha\right) \right)- \nonumber \\
   a_{2}^{2}\left(a_{1}^{2}+\frac{b^{2}}{4}-2a_{1}\frac{b}{2} \cos \left( \frac{\pi}{2}-\alpha\right)\right)=0 .
\end{gather}
%
%
After elementary modifications, the following equation is obtained:
\begin{eqnarray}
b=\frac{4a_{1}a_{2} s_\alpha}{a_{1}+a_{2}} .
\end{eqnarray}
Therefore, the scale is
\begin{gather}
 \frac{a}{b}=\frac{a_{1}+a_{2}}{\frac{4a_{1}a_{2}s_\alpha}{a_{1}+a_{2}}}=\frac{a_{1}^{2}+a_{2}^{2}+2a_{1}a_{2}}{4a_{1}a_{2}s_\alpha}= \nonumber \\ \frac{1}{4s_\alpha} \left(\frac{a_{1}}{a_{2}}+\frac{a_{2}}{a_{1}}\right) + \frac{1}{2s_\alpha}.
 \label{eq:ab_inhom}
\end{gather}
Due to the law of sines:
\begin{gather}
    \frac{a_{1}}{a_{2}}+\frac{a_{2}}{a_{1}}= 
    \frac{sin(\alpha+\beta)}{sin(\alpha-\beta)}+\frac{sin(\alpha-\beta)}{sin(\alpha+\beta)}= \nonumber \\
    \frac{s_\alpha c_\beta+c_\alpha s_\beta}{s_\alpha c_\beta-c_\alpha s_\beta}+\frac{s_\alpha c_\beta-c_\alpha s_\beta}{s_\alpha c_\beta+c_\alpha s_\beta}= \nonumber \\
  \frac{(s_\alpha c_\beta+c_\alpha s_\beta)^{2}+(s_\alpha c_\beta-c_\alpha s_\beta)^{2}}{s^{2}_\alpha c^{2}_\beta-c^{2}_\alpha s^{2}_\beta}=  \nonumber \\
  \frac{2s^{2}_\alpha c^{2}_\beta+2c^{2}_\alpha s^{2}_\beta}{s^{2}_\alpha c^{2}_\beta-c^{2}_\alpha s^{2}_\beta}  .
  \label{eq:a1a2}
\end{gather}
Hence, substituting Eq.~\ref{eq:a1a2} into Eq.~\ref{eq:ab_inhom}, the searched ratio:
\begin{gather}
    \frac{a}{b}=
    \frac{1}{4s_\alpha}\left(\frac{2s^{2}_\alpha c^{2}_\beta+2c^{2}_\alpha s^{2}_\beta}{s^{2}_\alpha c^{2}_\beta-c^{2}_\alpha s^{2}_\beta}\right)+\frac{1}{2s_\alpha}= \nonumber \\
    \frac{2s^{2}_\alpha c^{2}_\beta}{2s_\alpha(s^{2}_\alpha c^{2}_\beta-c^{2}_\alpha s^{2}_\beta)}= 
    \frac{s_\alpha c^{2}_\beta}{s^{2}_\alpha c^{2}_\beta-c^{2}_\alpha s^{2}_\beta}.
\end{gather}

\end{document}